\def\year{2019}\relax
\documentclass[letterpaper]{article} 
\usepackage{aaai19}  
\usepackage{times}  
\usepackage{helvet}  
\usepackage{courier}  
\usepackage{url}  
\usepackage{graphicx}  

\usepackage{amsmath}
\usepackage{amssymb}
\usepackage{subfigure}
\usepackage{algorithm,algorithmic}
\usepackage[usenames, dvipsnames]{color}

\usepackage{CJKutf8}

\usepackage{comment}

\begin{document}
%
\title{``Bilingual Expert" Can Find Translation Errors}
\author{Kai Fan\thanks{indicates equal contribution.}, Jiayi Wang$^*$, Bo Li$^*$, Fengming Zhou, Boxing Chen, Luo Si\\
\url{{k.fan,joanne.wjy,shiji.lb,zfm104435,boxing.cbx,luo.si}@alibaba-inc.com}\\
Alibaba Group Inc.\\
}
\maketitle
\begin{abstract}
The performances of machine translation (MT) systems are usually evaluated by the metric BLEU when the golden references are provided. 
However, in the case of model inference or production deployment, golden references are usually expensively available, such as human annotation with bilingual expertise. 
In order to address the issue of translation quality estimation (QE) without reference, we propose a general framework for automatic evaluation of the translation output for the QE task in the Conference on Statistical Machine Translation (WMT). 
We first build a conditional target language model with a novel bidirectional transformer, named \textbf{\textit{neural bilingual expert}} model, which is pre-trained on large parallel corpora for feature extraction. 
For QE inference, the bilingual expert model can simultaneously produce the joint latent representation between the source and the translation, and real-valued measurements of possible erroneous tokens based on the prior knowledge learned from parallel data. 
Subsequently, the features will further be fed into a simple Bi-LSTM predictive model for quality estimation. 
The experimental results show that our approach achieves the state-of-the-art performance in most public available datasets of WMT 2017/2018 QE task.
\end{abstract}

\section{Introduction}

\noindent The neural machine translation (NMT) in a sequence-to-sequence fashion, empowering an end-to-end learning approach for automatic translation system, has accomplished great success to potentially overcome many of the weaknesses of conventional phrase-based translation, and claimed being close to human parity for certain language pairs \cite{wu2016google,hassan2018achieving}. 
However, current MT systems are still not perfect to meet the real-world applications without human post-editing (a popular example is the Chinese to English translation test at Google online system \footnote{\scriptsize{This error appeared on 04/03/2018, and has been fixed.}}, which translated ``\begin{CJK}{UTF8}{gbsn}苹果比谷歌厉害\end{CJK}" to ``Apple is worse than Google", where the correct translation should be ``Apple is better than Google".). 
Apparently, additional error correction is needed for even such a simple translation output. 
A possible solution to take advantage of the existing MT technologies is to collaborate with human translators within a computer-assisted translation (CAT) \cite{barrachina2009statistical}. 
In such cases, translation quality estimation (QE) plays a critical role in CAT to reduce human efforts, thereby increasing productivity \cite{specia2011exploiting}. 
Either the global sentence quality score or the fine-grained word ``OK/BAD" tags can guide the CAT as an evidence to indicate whether a machine translation output requires further manual post-editing, or even which particular token needs special correction. 

One traditional direction for translation quality estimation is to formulate the sentence level score or word level tags prediction as a constraint regression or sequence labeling problem respectively \cite{bojar2017findings}. 
The classical baseline model is to use the QuEst++ \cite{specia-paetzold-scarton:2015:ACL-IJCNLP-2015-System-Demonstrations} with two modules: rule based feature extractor and scikit-learn \footnote{\scriptsize\url{http://scikit-learn.org/}} SVM algorithms. 
Similarly, the recent predictor-estimator model \cite{kim2017predictor} is a recurrent neural network (RNN) based feature extractor and quality estimation model, ranking first place at WMT 2017 QE. 

Another promising direction is to build a multi-task learning model to incorporate quality estimation task with automatic post-editing (APE) together \cite{hokamp2017ensembling,tan2017neural,chatterjee2018combining}, achieving the goal of CAT eventually.  
In this paper, we will first adopt the traditional single task framework to describe our model. 
In the experimental section, we also propose an extension to support multi-task learning for QE and APE simultaneously. 

However, the final prediction model for scoring or tagging is not the main contribution in our work. 
Since there are many publicly available bilingual corpora, we can readily build a conditional language model as a robust feature extractor. 
The high level joint latent representation of the source and the target in a parallel pair can hopefully capture the alignment or semantic information. 
In contrast, when a source and a low-quality machine translation are fed into the pre-trained language model, the distribution of latent features is very likely to be different from the one that grammatically correct target has. 
Intuitively, people can learn the foreign language from reading the correct translation to their native language. 
Gradually, they may acquire the ability to be aware of the abnormality, even when errors appear in a sentence have never seen before. 
Additionally, we design 4-dimensional token mis-matching features from the pre-trained model, measuring the difference between what the bilingual expert model will predict and the actual token of machine translation output. 

Particularly, we use the recent proposed self-attention mechanism and transformer neural networks \cite{vaswani2017attention} to build the conditional language model -- \textit{neural bilingual expert}. 
The model consists of the traditional transformer encoder for the source sentence and a novel bidirectional transformer decoder for the target sentence. 
It will be pre-trained on the large parallel corpus, and then produce high level features for the downstream quality estimation task. 
Constructing pre-trained word embedding on our designed language models has shown great improvement in many downstream NLP tasks. 
Both ELMO \cite{Peters:2018} and the OpenAI's transformer decoder trained for monolingual language model \cite{radfordimproving} are good illustrations. 
Bidirectional attention mechanism was mainly proposed to achieve success in machine reading comprehension, such BiDAF \cite{seo2016bidirectional} and \cite{shen2018bi}. 
However, all of them are used for monolingual training without involving other conditional language. 

The conditional language model can play the role of automatic post-editing as well. 
Since shifts were not annotated as word order errors (but rather as deletions and insertions) to avoid introducing noise in the annotation, missing tokens in the machine translations, as indicated by the TER tool \cite{snover2006study}, are annotated as follows: after each token in the sentence and at sentence start, a gap tag is placed. 
In this situation, we can use the same network structure of conditional language model to enable the gap prediction (insertions) for missing token of translation output conditional on the source sentence. 
Using the deletion operation in word level tagging (by adding class ``D" rather than ``OK/BAD"), we are literally trying to predict post-editing. 

This paper makes the following main contributions: 
i) we propose a novel approach with bidirectional transformer for building a conditional language model and pre-train it on available large bilingual corpora, which can further be used as automatic post-editing model. 
ii) we address the importance of the 4-dimensional mis-matching features, and in the experiments, with only these features, our approach can still achieve comparable results with No. 1 system in WMT 2017 QE task.
iii) we develop a differentiable word-level quality estimation model to support data preprocessing with byte-pair-encoding (BPE) tokenization, bridging the gap between words and BPE tokens. 
iv) extensive experiments on real-world datasets (e.g., IT and pharmacy domain corpora) demonstrate our method is effective and achieve the state-of-the-art performance in most tasks.

\section{Background}

\paragraph{Quality Estimation for Machine Translation}
Given the bilingual corpus, from the statistical view we can formulate the machine translation system as $p(\mathbf{t}|\mathbf{s})=p(\mathbf{t}|\mathbf{z})p(\mathbf{z}|\mathbf{s})$, where $\mathbf{s}$ represents the tokens sequence of source sentence, $\mathbf{t}$ for target sentence, and $\mathbf{z}$ is the latent variable to represent the encoded source sentence. 
Therefore, $p(\mathbf{z}|\mathbf{s})$ and $p(\mathbf{t}|\mathbf{z})$ can be practically considered as the encoder and decoder.
In the quality estimation task of machine translation, the machine translation system is agnostic and the training dataset is given in the format of triplet $(\mathbf{s}, \mathbf{m}, \mathbf{t})$, where $\mathbf{m}$ is the translation output from the unknown machine translation system with the input $\mathbf{s}$, and $\mathbf{t}$ represents the human post-edited sentence based on $\mathbf{s}$ and $\mathbf{m}$. 
Notice we abuse using notation $\mathbf{t}$ to refer both golden reference and human post-edited sentence. 

In general, the quality of $\mathbf{m}$ can be evaluated either in the global sentence level or the fine-grained word level. 
The sentence level score is calculated by the percentage of edits needed to fix for $\mathbf{m}$, denoted as HTER. 
The word level evaluation is framed as the sequential binary classification problem to distinguish between `OK' and `BAD' for each token in translation output. 
Particularly, the binary word-level labels are generated by using the alignments provided by the TER tool \cite{snover2006study} 
between $\mathbf{m}$ and $\mathbf{t}$.
Notice the sentence HTER and word labels can also deterministically be calculated by the TER tool when $\mathbf{m}$ and $\mathbf{t}$ are both present. 
However, in inference only the source sentence $\mathbf{s}$ and machine translation $\mathbf{m}$ are available, thus essentially requiring an automatic method for quality estimation of machine translation output at run-time, without relying on any reference.

We can assume that the training data contains the tuple $(\mathbf{s}, \mathbf{m}, \mathbf{t}, h, \mathbf{y})$, where $h$ is a scalar to represent HTER, and $\mathbf{y}$ is a binary vector to indicate the `OK/BAD' labels of machine translation output. 
Considering the inference scenario, our task is to learn a regression model $p(h|\mathbf{s}, \mathbf{m})$ and a sequence labeling model $p(\mathbf{y}|\mathbf{s}, \mathbf{m})$.

\begin{figure*}[ht]
\centering
\includegraphics[width=\textwidth]{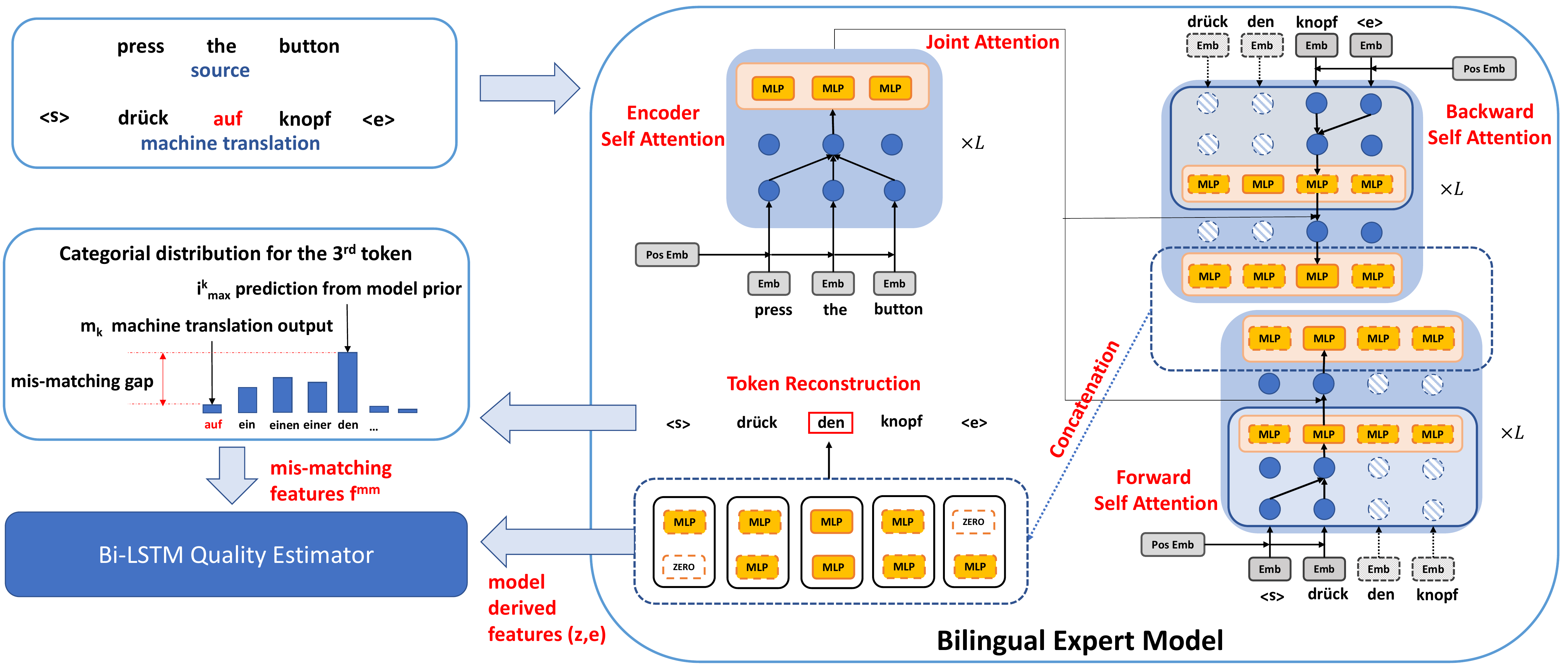}
\caption{Right: Bilingual Expert Model. The encoder is basically identical to the transformer NMT. The forward and backward self-attentions mimic the structure of bidirectional RNN, implemented by the left to right and right to left masked softmax respectively. \textit{Notice} that some detailed network structures, like skip-connection and layer normalization, are omitted for clarity. Left: Quality Estimation Model. Two features are derived from the pre-trained bilingual expert model.}
\label{fig:bi_trans}
\end{figure*}

\section{Methodology}
\subsection{Bilingual Expert Model}

In this section, we will first highlight how to train a neural bilingual expert model with a parallel corpus including $(\mathbf{s},\mathbf{t})$ pairs. 
By default of QE task, the machine translation system $p(\mathbf{t}|\mathbf{z})p(\mathbf{z}|\mathbf{s})$ is unknown, but in representation learning we are usually interested in the latent variable $\mathbf{z}$, whose posterior may contain the deep semantic information between the source and the target languages, and be beneficial to many downstream tasks \cite{hill2016learning}. 
According to the Bayes rule, we can write the posterior distribution of the latent variable as,
\begin{equation}
p(\mathbf{z}|\mathbf{t}, \mathbf{s}) = \frac{p(\mathbf{t}|\mathbf{z})p(\mathbf{z}|\mathbf{s})}{p(\mathbf{t}|\mathbf{s})}
\end{equation}
where the integral $p(\mathbf{t}|\mathbf{s})=\int p(\mathbf{t}|\mathbf{z})p(\mathbf{z}|\mathbf{s}) \mathrm{d}\mathbf{z}$ is usually intractable. 
Instead of exact inference, we propose a variational distribution $q(\mathbf{z}|\mathbf{t}, \mathbf{s})$ to approximate true posterior by minimizing exclusive Kullback-Leibler (KL) divergence. 
\begin{equation}
\min D_{KL}(q(\mathbf{z}|\mathbf{t}, \mathbf{s})\|p(\mathbf{z}|\mathbf{t}, \mathbf{s}))
\end{equation}
Rather than optimizing the objective function above, we can equivalently maximize the following one,
\begin{equation}\label{eq:elbo}
\max \mathbb{E}_{q(\mathbf{z}|\mathbf{t}, \mathbf{s})}[p(\mathbf{t}|\mathbf{z})] - D_{KL}(q(\mathbf{z}|\mathbf{t}, \mathbf{s})\|p(\mathbf{z}|\mathbf{s})) .
\end{equation}
A nice property of the new objective is that it is unnecessary to parameterize or estimate the implicit machine translation model $p(\mathbf{t}|\mathbf{s})$. 
The first expectation term in (\ref{eq:elbo}) can be readily considered as a conditional auto-encoder system if we use one sample Monte Carlo integration during optimization, and the second KL term can be analytically computed if we practically set the prior $p(\mathbf{z}|\mathbf{s})$ as standard Gaussian distribution, playing as a model regularization for latent variables. 
Furthermore, if we omit the conditional information $\mathbf{s}$, the objective exactly reduces to amortized variational inference or variational auto-encoders (VAE) framework \cite{kingma2013auto}. 
In analogous to most VAE models, the expected log-likelihood is commonly approximated by a practical surrogated term,
\begin{equation}
\mathbb{E}_{q(\mathbf{z}|\mathbf{t}, \mathbf{s})}[p(\mathbf{t}|\mathbf{z})] \approx p(\mathbf{t}|\tilde{\mathbf{z}}), ~~\tilde{\mathbf{z}}\sim q(\mathbf{z}|\mathbf{t}, \mathbf{s})
\end{equation}
Next, we will show the details of constructing the other two probability distributions appeared in (\ref{eq:elbo}) with self-attention based transformer neural networks. 

\subsection{Bidirectional Transformer}
Transformer \cite{vaswani2017attention} is based solely on attention mechanisms, dispensing with recurrence and convolution, becoming the state-of-the-art NMT model in most machine translation competitions. 
\citeauthor{vaswani2017attention} claims that self-attention mechanism has several advantages: first, its gating or multiplication enables crisp error propagation; second, it can replace sequence-aligned recurrence entirely; third, from the implementation perspective, it is trivial to be parallelized during training. 
When we design the bidirectional transformer, we are trying to keep the three properties remained in our model.  

The overall model architecture of bidirectional transformer is illustrated in the right block of Figure~\ref{fig:bi_trans}. 
There are three modules in total, self-attention encoder for the source sentence, forward and backward self-attention encoders for target sentence, and the reconstructor for the target sentence, where the first two modules represent the proposed posterior approximation $q(\mathbf{z}|\mathbf{s},\mathbf{t})$ and the third reconstruction process corresponds to $p(\mathbf{t}|\mathbf{z})$. 
To make the inference efficient, we explicitly assume the conditional independence with the following factorization,
\begin{align}
p(\mathbf{t}|\mathbf{z}) &= \prod_{k}p(t_k|\overrightarrow{\mathbf{z}_k},\overleftarrow{\mathbf{z}_k}) \label{eq:p_t} \\
q(\mathbf{z}|\mathbf{s},\mathbf{t}) &= \prod_{k} q(\overrightarrow{\mathbf{z}_k}|\mathbf{s},\mathbf{t}_{<k})q(\overleftarrow{\mathbf{z}_k}|\mathbf{s},\mathbf{t}_{>k}) \label{eq:q_z}
\end{align}
where the bidirectional latent variable $\mathbf{z}$ includes all $\{\overrightarrow{\mathbf{z}_k},\overleftarrow{\mathbf{z}_k}\}$. 
Note that our factorization is different from ELMO \cite{Peters:2018}, where they use a finer grained form $\prod_{k}p(t_k|\overrightarrow{\mathbf{z}_k})p(t_k|\overleftarrow{\mathbf{z}_k})$ but with the shared parameters between forward and backward reconstruction $p(t_k|\cdot)$. 

Latent variables $\overrightarrow{\mathbf{z}_k},\overleftarrow{\mathbf{z}_k}$ are sampled from $q(\overrightarrow{\mathbf{z}_k}|\mathbf{s},\mathbf{t}_{<k})$ and $q(\overleftarrow{\mathbf{z}_k}|\mathbf{s},\mathbf{t}_{>k})$ respectively, assuming to follow the Gaussian distribution, e.g., $q(\cdot|\cdot) \sim \mathcal{N}(\mu(\mathbf{s},\mathbf{t}), \sigma^2\mathbf{I})$. 
Meanwhile, the mean $\mu(\mathbf{s},\mathbf{t})$ is learned in an amortized way, i.e., every single pair $\mathbf{s},\mathbf{t}$ will generate their own mean via the shared neural network model. 
By fixing $\sigma$ as a hyper-parameter, we can efficiently implement the stochastic layer as the deterministic one via dropout training with additive Gaussian noise \cite{srivastava2014dropout}. 
The stochastic layer can increase the uncertainty of the latent representation, potentially preventing overfitting. 
In practice, a small $\sigma$ is recommended. 
Notice that we didn't follow the NMT parlance to call our bidirectional self-attention transformer as ``decoder",  since it is not actually a generative model during inference. 

\subsection{Model Derived Features}

Once the bilingual expert model has been fully trained on large parallel corpora, we can reasonably assume the model will predict higher likelihood for the correct target token, given the source and other context of the target, if only very few tokens are incorrect. 
Therefore, we will use the prior knowledge learned by bilingual expert to extract the features for subsequent translation error prediction. 
Basically, we will first design the sequential (token-wise) model derived features based upon the pre-trained model with $(\mathbf{s},\mathbf{m})$ pair as input.
The latent representation $\mathbf{z}_k = \text{Concat}(\overrightarrow{\mathbf{z}_k}, \overleftarrow{\mathbf{z}_k})$ should naturally be the high level features. 
As we discussed previously, the entire latent variable $\mathbf{z}$ should generally summarize the information of the source and the target. 
In Equation (\ref{eq:q_z}), the distribution of $\mathbf{z}_k$ is deliberately defined to contain the information from the source and the context around the $k$-th token in the target.  
We see this by observing the computational graph in the right panel of Figure~\ref{fig:bi_trans}, e.g., the token ``den" of target is desired to predict, but only the information of the source and all the other tokens in the target will be propagated to the final layer for prediction. 
It will be reasonably beneficial to our manually extracted mis-matching features introduced later.

In ELMO \cite{Peters:2018}, the token embedding is also used as one linear component to compute the final feature. 
However, in our case that translation output is fed into the model, it is not guaranteed that every single token is correct. 
Therefore, we design a different token embedding feature following the rationale of subtle information flow within latent variable $\mathbf{z}_k$. 
In fact, we use the embedding concatenation of two neighbor tokens $\text{Concat}(\mathbf{e}_{t_{k-1}}, \mathbf{e}_{t_{k+1}})$. 
Since the possibly erroneous translation may mislead the model in the downstream quality estimation task, we did not extract any information from current token $t_k$. 
More importantly, the correct syntax representation of the token which is supposed to be translated should come from the source sentence, which has been encoded into $\mathbf{z}$ via joint attention.

\subsection{Mis-matching Features}

Besides the proposed model derived features that are exactly the nodes within the computational graph of the bidirectional transformer, we intuitively found another type of crucial features that can directly measure how the prior knowledge from the well-trained bilingual expert model is different from the translation. 
To make it concrete, $p(t_k|\cdot)$ follows the categorical distribution with the number of classes equal to the vocabulary size. 
Since we pre-train the bilingual expert model on parallel corpara, the objective (\ref{eq:elbo}) is theoretically to maximize the likelihood of each $p(t_k|\cdot)$, which achieves its maximum when $t_k$ is ground truth. 
Intuitively, we should have $p(m_k|\cdot) \leq p(t_k|\cdot)$ for optimal model if $m_k \neq t_k$, illustrated in top-left block of Figure~\ref{fig:bi_trans}. 
Following this intuition, we propose the mis-matching features. 

Suppose $\mathbf{l}_k$ is the logits vector before applying the softmax operation, i.e. $p(t_k|\cdot) \sim \text{Categorical}(softmax(\mathbf{l}_k))$, thus we can define the 4-dimensional mis-matching features as the following vector,
\begin{equation}
\mathbf{f}_k^{mm} = (\mathbf{l}_{k, m_k}, \mathbf{l}_{k, i^k_{\max}}, \mathbf{l}_{k, m_k}-\mathbf{l}_{k, i^k_{\max}}, \mathbb{I}_{m_k\neq i^k_{\max}}) 
\end{equation}
where $m_k$ represents the vocabulary id of the $k$-th token in translation output,  $i^k_{\max} = \arg\max_i \mathbf{l}_k$ is the id that the bilingual expert predicts, and $\mathbb{I}$ is indicator function. 
Therefore, these four values will directly reflect the differences or errors. 
Apparently, if the machine translation coincides with the bilingual expert prediction, the first 2 elements of $\mathbf{f}_k^{mm}$ should be identical and the last two elements, representing soft and hard differences, should be both 0. 
We empirically found the quality estimation model can achieve comparable result even with the mis-matching features alone. 

\begin{algorithm}[t]
\caption{Translation Quality Estimation with Bi-Transformer and Bi-LSTM}
\begin{algorithmic}[1]
\small
\REQUIRE QE training data $(\mathbf{s}, \mathbf{m}, \mathbf{t}, h, \mathbf{y})_{1:M}$, QE inference data $(\mathbf{s},\mathbf{m})$, and parallel corpus $(\mathbf{s},\mathbf{t})_{1:N}$.
\STATE Combine the parallel corpus with 10 copies of QE training parallel corpus $C = (\mathbf{s}_n,\mathbf{t}_n)_{n=1}^N \bigcup 10 \times (\mathbf{s}_m, \mathbf{t}_m)_{m=1}^M$
\STATE Pre-train bilingual expert model via the bidirectional transformer on the combined corpus $C$.
\STATE Extract features $\mathbf{f}_k = \text{Concat}(\overrightarrow{\mathbf{z}_k}, \overleftarrow{\mathbf{z}_k}, \mathbf{e}_{t_{k-1}}, \mathbf{e}_{t_{k+1}}, \mathbf{f}_k^{mm})$ for QE training data $(\mathbf{s}, \mathbf{m})$.
\STATE Train Bi-LSTM model via objectives (\ref{eq:sent_loss})(\ref{eq:word_loss}).
\RETURN Predict $h, \mathbf{y}$ for QE inference data
\end{algorithmic} 
\label{alg:double_bi}
\end{algorithm}

\begin{table*}[t]
\setlength{\tabcolsep}{2pt}
\small
\centering
\begin{tabular}{c|c|c|c|c|c|c|c|c|c|c}
 & \multicolumn{5}{|c|}{test 2017 en-de} & \multicolumn{5}{|c}{test 2017 de-en} \\
 \cline{2-11}
Method & \textbf{Pearson's} $\uparrow$& MAE $\downarrow$& RMSE $\downarrow$& \textbf{Spearman's}  $\uparrow$ & DeltaAvg $\uparrow$ & \textbf{Pearson's} $\uparrow$& MAE $\downarrow$& RMSE $\downarrow$& \textbf{Spearman's} $\uparrow$ & DeltaAvg $\uparrow$ \\
\hline
\hline
Baseline & 0.3970 & 0.1360 & 0.1750 & 0.4250 & 0.0745 & 0.4410 & 0.1280 & 0.1750 & 0.4500 & 0.0681\\
Unbabel & 0.6410 & 0.1280 & 0.1690 & 0.6520 & 0.1136 & 0.6260 & 0.1210 & 0.1790 & 0.6100 & 0.9740 \\
POSTECH Single & 0.6599 & 0.1057 & 0.1450 & 0.6914 & 0.1188 & 0.6985 & 0.0952 & 0.1461 & 0.6408 & 0.1039\\
\hline
Ours Single (MD+MM) & \textbf{0.6837} & 0.1001 & 0.1441 & \textbf{0.7091} & 0.1200 & \textbf{0.7099} & 0.0927 & 0.1394 & \textbf{0.6424} & 0.1018 \\
w/o MM & 0.6763 & 0.1015 & 0.1466 & 0.7009 & 0.1182 & 0.7063 & 0.0947 & 0.1410  & 0.6212 & 0.1005\\
w/o MD & 0.6408 & 0.1074 & 0.1478 & 0.6630 & 0.1101 & 0.6726 & 0.1089 & 0.1545 & 0.6334  & 0.0961\\
\hline
\hline
POSTECH Ensemble & 0.6954 & 0.1019 & 0.1371 & 0.7253 & 0.1232 &  0.7280 & 0.0911 & 0.1332 & 0.6542 & 0.1064 \\
Ours Ensemble & \textbf{0.7159} &  0.0965 & 0.1384 & \textbf{0.7402} & 0.1247 & \textbf{0.7338} & 0.0882 & 0.1333 & \textbf{0.6700} & 0.1050 \\
\hline
\end{tabular}
\caption{Results of sentence level QE on WMT 2017. MD: model derived features. MM: mis-matching features.}
\label{tab:sent2017}
\end{table*}

\subsection{Bi-LSTM Quality Estimation}
To this end, we have the model derived and manually designed sequential features, each time stamp of which is corresponding to a fixed size vector. 
Our quality estimation task is built upon the bidirectional LSTM \cite{graves2005framewise} model, being widely used for sequence classification or sequence tagging problems. 
In sequence tagging, \citeauthor{huang2015bidirectional} proposed a variant of Bi-LSTM with one Conditional Random Field (CRF) layer (Bi-LSTM-CRF). 
We empirically found that the extra CRF layer did not show any significant improvement over vanilla Bi-LSTM, which we simply adopted. 
Another natural question is whether the traditional encoder self-attention or our proposed forward/backward self-attention can be an alternative to the Bi-LSTM. 
We empirically found the results with self-attention module become even worse, and we suspect the scarcity of labelled quality estimation data, being incomparable to the sufficient parallel corpus, is the main reason.

We concatenate all sequential features along the depth direction to obtain a single vector, denoted as $\{\mathbf{f}_k\}_{k=1}^T$, where $T$ is the number of tokens in $\mathbf{m}$. 
Therefore, the sentence level score HTER prediction can be formulated as a regression problem (\ref{eq:sent_loss}), and the word error prediction is a sequence labeling problem (\ref{eq:word_loss}),
\begin{align}
\overrightarrow{\mathbf{h}_{1:T}}, \overleftarrow{\mathbf{h}_{1:T}} &= \text{Bi-LSTM}(\{\mathbf{f}_k\}_{k=1}^T) \\
\arg\min & \left\|h - \text{sigmoid}\left(\mathbf{w}^\top [\overrightarrow{\mathbf{h}_T}, \overleftarrow{\mathbf{h}_T}]\right) \right\|_2^2 \label{eq:sent_loss} \\
\arg\min & \sum_{k=1}^T \text{XENT}(y_k, \mathbf{W}[\overrightarrow{\mathbf{h}_k}, \overleftarrow{\mathbf{h}_k}]) \label{eq:word_loss}
\end{align}
where $\mathbf{w}$ is a vector, $\mathbf{W}$ is a matrix, $y_k$ is the error label for the $k$-th token of translation output, and XENT is the cross entropy loss (with logits). 
Notice HTER $h$ is a real value within interval $[0,1]$, we apply a squash function ``sigmoid" for rescaling in the regression model. 
Since the HTER is a global score for the entire sentence, we use the hidden states of the last time stamp in the forward/backward LSTMs as the regression signals. 
Actually, we can train the two losses together in a multi-task setting. 
In summary, we describe the outline of our proposed approach in Algorithm~\ref{alg:double_bi}.  

\section{Experiments}


\subsection{Setting Description}
The data resources that we used for training the neural Bilingual Expert model are mainly from WMT\footnote{\scriptsize\url{http://www.statmt.org/wmt18/}}: (i) parallel corpora released for the WMT17/18 News Machine Translation Task, (ii) UFAL Medical Corpus and Khresmoi development data released for the WMT17/18 Biomedical Translation Task, (iii) src-pe pairs for the WMT17/18 QE Task. To ensure the quality of the corpora, we filtered the source and target sentence with length $\leq$70 and the length ratio between 1/3 to 3, thus resulting roughly 9 million (2017) and 25 million (2018) parallel sentences pairs for both English$\leftrightarrow$German directions. 
We mainly tried word tokenization for the corpus in the WMT17 QE task, where the word tokenization naturally fits the word level QE task. 
For WMT18, we applied byte-pair-encoding (BPE) \cite{sennrich2016neural} tokenization to reduce the number of unknown tokens. 
However, there exists the discrepancy between word token tagging prediction and BPE tokenization, and we will present how to bridge the gap in the next section. 
We also test our model on the CWMT 2018 Chinese English sentence QE task\footnote{\scriptsize\url{http://nlp.nju.edu.cn/cwmt2018/guidelines.html}}. 
Since the two languages are unrelated, we tokenize them separately.

The number of layers in the bidirectional transformer for each module is 2, and the number of hidden units for feedforward sub-layer is 512. 
We use the 8-head self-attention in practice, since the single one is just a weighted average of previous layers. 
The bilingual expert model is trained on 8 Nvidia P-100 GPUs for about 3 days until convergence. 
For translation QE model, we use only one layer Bi-LSTM, and it is trained on a single GPU. 

We evaluate our algorithm on the testing data of WMT 2017/2018, and development data of CWMT 2018.
Notice for the QE task of WMT 2017, it is forbidden to use any data from 2018, since the training data of 2018 includes some testing data of 2017. 
The same setting applies to all following experiments. 
For fair comparison, we tuned all the hyper-parameters of our model on the development data, and reported the corresponding results for the testing data. 

\subsection{Sentence Level Scoring And Ranking} 

\begin{table}[t]
\setlength{\tabcolsep}{2pt}
\small
\centering
\begin{tabular}{c|c|c|c|c}
 & \textbf{Pearson's} $\uparrow$& MAE $\downarrow$& RMSE $\downarrow$& \textbf{Spearman's} $\uparrow$ \\
\cline{2-5}
Method & \multicolumn{4}{|c}{test 2018 en-de} \\
\hline
Baseline & 0.3653 & 0.1402 & 0.1772 & 0.3809 \\
UNQE & 0.7000 & 0.0962 & 0.1382 & 0.7244\\
\hline
Ours Ensemble & \textbf{0.7308} & 0.0953 & 0.1383 & \textbf{0.7470} \\
\hline
Method & \multicolumn{4}{|c}{test 2018 de-en} \\
\hline
Baseline & 0.3323 & 0.1508 & 0.1928 & 0.3247 \\
UNQE & \textbf{0.7667} & 0.0945 & 0.1315 & 0.7261 \\
\hline
Ours Ensemble & 0.7631 & 0.0962 & 0.1328 & \textbf{0.7318} \\
\hline
\end{tabular}
\caption{Results of sentence level QE on WMT 2018}
\label{tab:sent2018}
\end{table}

\begin{table}[t]
\setlength{\tabcolsep}{2pt}
\small
\centering
\begin{tabular}{c|c|c|c}
System & Used Bi-Corpus & Ch->En & En->Ch  \\
\hline
CWMT 1st ranked (Ensemble) & CWMT 8m + 8m BT & 0.465 & 0.405 \\
Our Model 1 (Single) & WMT 25m + 25m BT & 0.612 & 0.620 \\
Our Model 2 (Single) & CWMT 8m & 0.564 & 0.588 \\
\hline
\end{tabular}
\caption{Pearson's coefficient of CWMT 2018 QE}
\label{tab:sent2018_cwmt}
\end{table}


\begin{table}[h]
\small
\centering
\begin{tabular}{c|c|c|c}
 & F1-BAD & F1-OK & \textbf{F1-Multi} \\
\cline{2-4}
Method & \multicolumn{3}{|c}{test 2017 en-de} \\
\hline
Baseline & 0.407 & 0.886 & 0.361 \\
DCU  & 0.614 & 0.910 & 0.559 \\
Unbabel  & 0.625 & 0.906 & 0.566 \\
POSTECH Ensemble & 0.628 & 0.904 & 0.568 \\
\hline
Ours Single (MM + MD) & 0.6410 & 0.9083 & \textbf{0.5826} \\
\hline
Method & \multicolumn{3}{|c}{test 2017 de-en} \\
\hline
Baseline & 0.365 & 0.939 & 0.342 \\
POSTECH Single & 0.552 & 0.936 & 0.516 \\
Unbabel  & 0.562 & 0.941 & 0.529 \\
POSTECH Ensemble & 0.569 & 0.940 & 0.535 \\
\hline
Ours Single (MM + MD) & 0.5816 & 0.9470 & \textbf{0.5507} \\
\hline
\hline
Method & \multicolumn{3}{|c}{test 2018 en-de SMT} \\
\hline
Baseline & 0.4115 & 0.8821 & 0.3630 \\
Conv64 & 0.4768 & 0.8166 & 0.3894 \\
SHEF-PT  & 0.5080 & 0.8460 & 0.4298 \\
\hline
Ours Ensemble & 0.6616 & 0.9168 & \textbf{0.6066} \\
\hline
Method & \multicolumn{3}{|c}{test 2018 en-de NMT} \\
\hline
Baseline & 0.1973 & 0.9184 & 0.1812 \\
Conv64 & 0.3573 & 0.8520 & 0.3044 \\
SHEF-PT  & 0.3353 & 0.8691 & 0.2914 \\
\hline
Ours Ensemble & 0.4750 & 0.9152 & \textbf{0.4347} \\
\hline
Method & \multicolumn{3}{|c}{test 2018 de-en SMT} \\
\hline
Baseline & 0.4850 & 0.9015 & 0.4373 \\
Conv64 & 0.4948 & 0.8474 & 0.4193 \\
SHEF-PT  & 0.4853 & 0.8741 & 0.4242 \\
\hline
Ours Ensemble & 0.6475 & 0.9162 & \textbf{0.5932} \\
\hline
\end{tabular}
\caption{Results of word level QE on WMT 2017/2018}
\label{tab:word201718}
\end{table}


The sentence level results of WMT 2017 are listed in Table~\ref{tab:sent2017}. 
We mainly compared our single model with the two algorithms \cite{kim2017predictor,martins2017unbabel}, ranking top 3 in the WMT 2017 finalist. 
\textbf{Unbabel} is combination of a feature-rich sequential linear model with a neural network. 
\textbf{POSTECH} is a predictor-estimator model with all Bi-GRU modules. 
\textbf{Baseline} is the official provided system. 
The primary metrics of sentence level task are Pearson's correlation and Spearman's rank correlation of the entire testing data. 
Alternatively, mean average error (MAE), root mean squared error (RMSE), or the average of delta values (DeltaAvg) can also measure the performance of overall predictions, but not be a ranking reference in the QE task.
For both single and ensemble model comparisons, our algorithm can outperform all other systems for the two primary metrics. 
The ranking results are generated by the predicted HTER scores. 
In addition, we also analyze the importance of model derived features (MD) and mis-matching features (MM) the ablation study. 
With 4-dimensional mis-matching features alone, the model can still achieve comparable or better performance than the second single system last year. 
It demonstrates that the low dimensional features can provide a strong prediction signal as well.

We also report the result on unrelated language pair, Chinese and English, as shown in Table~\ref{tab:sent2018_cwmt}, where BT means back-translation. 
Our single model without back-translation has outperformed the best system in the competition. 

\subsection{Word Level For Word Tagging}
The metric of word level is evaluated in terms of classification performance via the multiplication of F1-scores for the `OK' and `BAD' classes against the true labels. 
For the binary classification, we tuned the threshold of the classifier on the development data and applied to the test data. 
The overall results are shown at Table~\ref{tab:word201718}. 
The baseline is provided by the offical WMT organizers, and the system is trained by CRFSuite toolkit with passive-aggressive algorithm \cite{CRFsuite}. 
We also compared the top 3 algorithms in WMT17 QE task, POSTECH \cite{kim2017predictor}, Unbabel \cite{martins2017unbabel}, and DCU \cite{martins2017pushing}. 
\textbf{DCU} is a stacked neural model by exploiting synergies between the related tasks of word-level quality estimation and automatic post-editing. 
In the primary metric F1-Multi, our algorithm of the single model outperforms all other models, including the best  ensemble system in WMT17. 
In WMT18 word level QE task, our approach exceeds all other algorithms with significant better numbers. 

The higher value of single F1-OK or F1-BAD cannot reflect the robustness of the algorithm, since it may result in lower F1 of another metric. 
Though we presented the F1-OK and F1-BAD, it is not a valid metric to QE task. 
However, by comparing them, we can conclude that all algorithms tend to classify the word tag as OK in general, since the true labels are very imbalanced. 
This is the reason why we use the threshold tuning strategy to finalize our classifier.


\subsection{Word Level For Gap Tagging}

\begin{table*}[h]
\setlength{\tabcolsep}{5pt}
\scriptsize
\begin{tabular}{c|c|c|c}
Method & F1-BAD & F1-OK & \textbf{F1-Multi} \\
\hline
UAlacante SBI & 0.1997 & 0.9444 & 0.1886 \\
SHEF-bRNN  & 0.2710 & 0.9552 & 0.2589 \\
SHEF-PT  & 0.2937 & 0.9618 & 0.2824 \\
\hline
Ours Ensemble & 0.5109 & 0.9783 & \textbf{0.4999}\\
\hline
\end{tabular}
\begin{tabular}{|c|l|}
\hline
MT & \colorbox{BurntOrange}{wählen} sie im bedienfeld " profile " des dialogfelds " preflight " auf die schaltfläche " \colorbox{BurntOrange}{längsschnitte} auswählen . " \\
APE & klicken sie im bedienfeld " profile " des dialogfelds " preflight " auf die schaltfläche " profile auswählen . " \\
PE & klicken sie im bedienfeld " profile " des dialogfelds " preflight " auf die schaltfläche " profile auswählen . " \\
\hline
MT &  das \colorbox{BurntOrange}{teilen} von komplexen symbolen und große textblöcke kann viel zeit in anspruch nehmen . \\
APE & das \colorbox{BurntOrange}{trennen} von komplexen symbolen und großen textblöcke kann viel zeit in anspruch nehmen . \\
PE &  das aufteilen von komplexen symbolen und großen textblöcke kann viel zeit in anspruch nehmen . \\
\hline
MT &  sie müssen nicht auf den ersten punkt , um das polygon zu schließen . \\
APE & sie müssen nicht auf den ersten punkt \colorbox{Yellow}{klicken} , um das polygon zu schließen . \\
PE &  sie müssen nicht auf den ersten punkt klicken , um das polygon zu schließen . \\
\hline
MT & sie können bis zu vier zeichen . \\
APE & sie können bis zu vier zeichen \colorbox{Yellow}{eingeben} . \\
PE &  sie können bis zu vier zeichen eingeben . \\
\hline
MT &  die standardmaßeinheit in illustrator \colorbox{BurntOrange}{beträgt punkte} ( ein punkt entspricht .3528 \colorbox{BurntOrange}{millimeter} ) . \\
APE & die standardmaßeinheit in illustrator ist punkt ( ein punkt entspricht .3528 \colorbox{BurntOrange}{millimeter} ) . \\
PE &  die standardmaßeinheit in illustrator ist punkt ( ein punkt entspricht .3528 millimetern ) . \\
\hline
\end{tabular}
\caption{Left Table: result of word level for gap prediction on WMT2018 En-De. Right Table: neural bilingual model with gap prediction expertise. In the shown examples, orange word means error translation, and yellow word means missing word. MT: machine translation; APE: automatic post-editing; PE: human post-editing.}
\label{tab:apegap}
\end{table*}

\begin{figure*}[t]
\centering
\subfigure[Segmentation matrix]{
\includegraphics[width=0.25\textwidth]{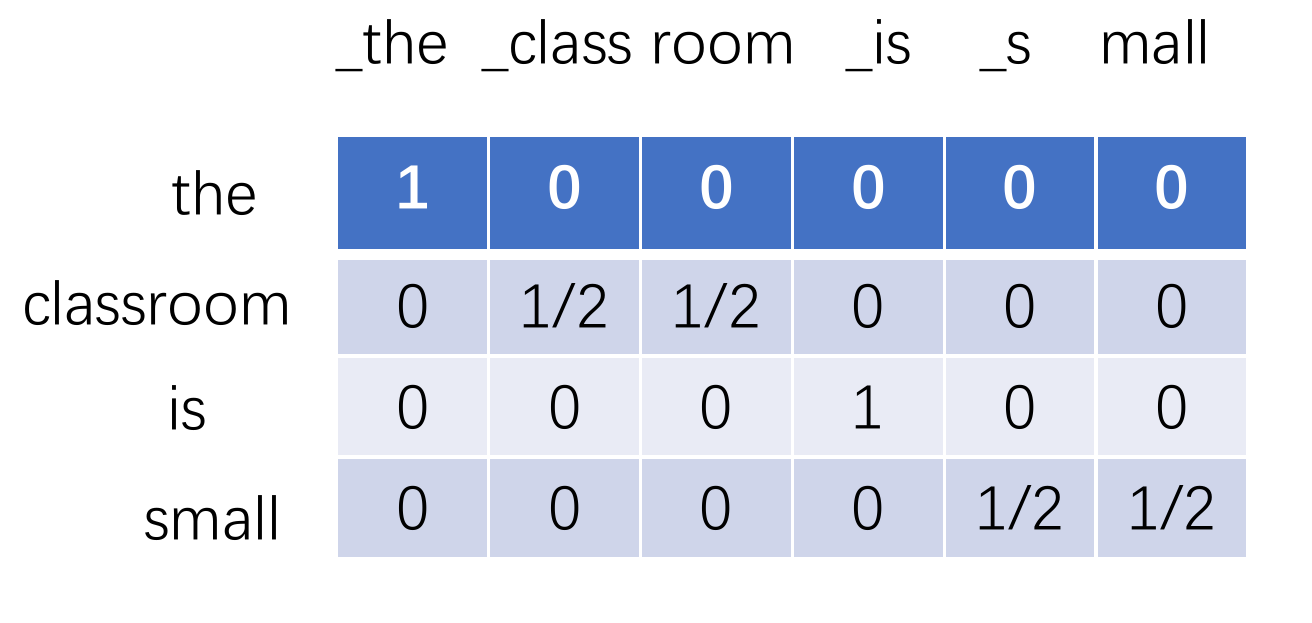}
}
\subfigure[Sentence Level]{
\includegraphics[width=0.33\textwidth]{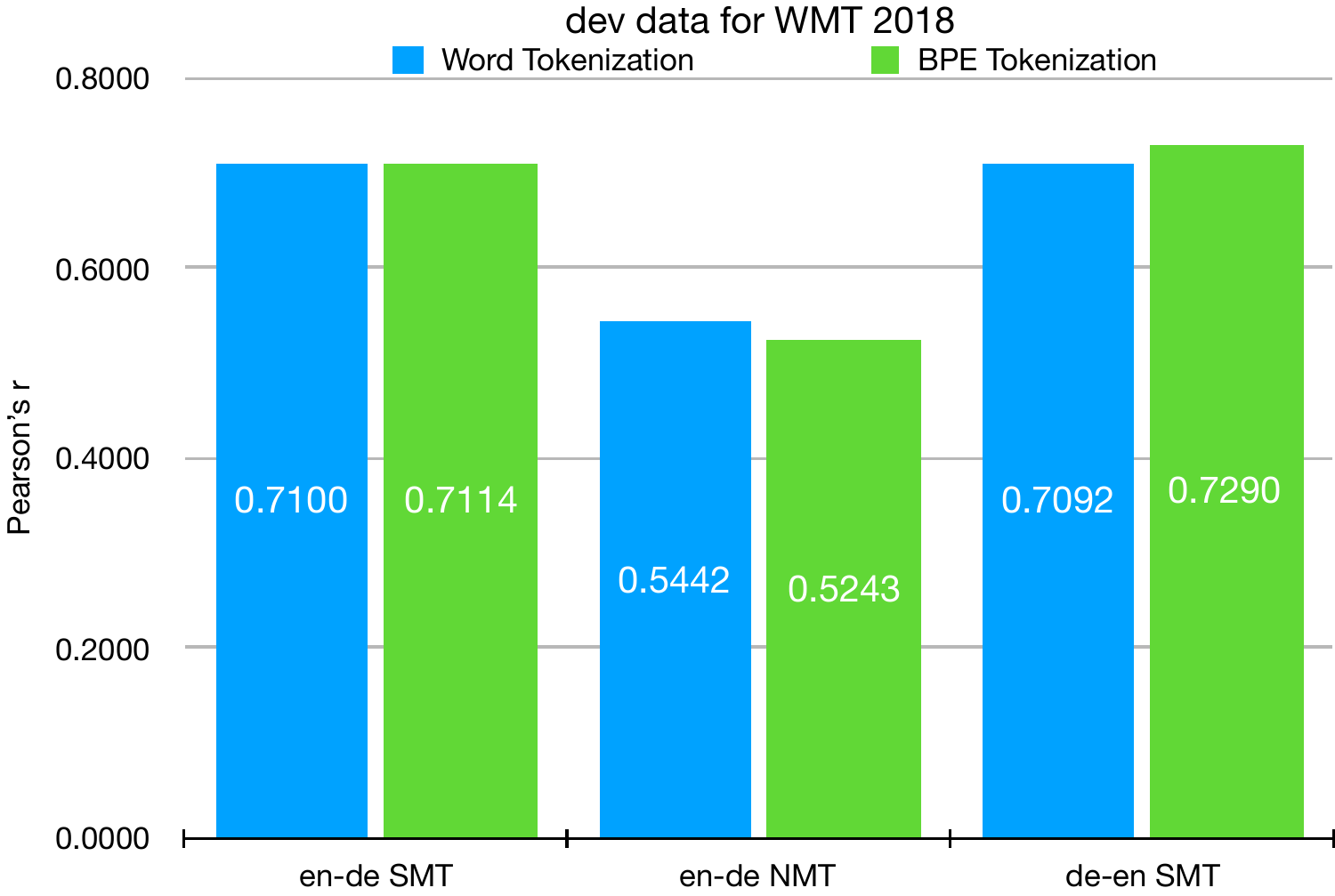}
}
\subfigure[Word Level]{
\includegraphics[width=0.33\textwidth]{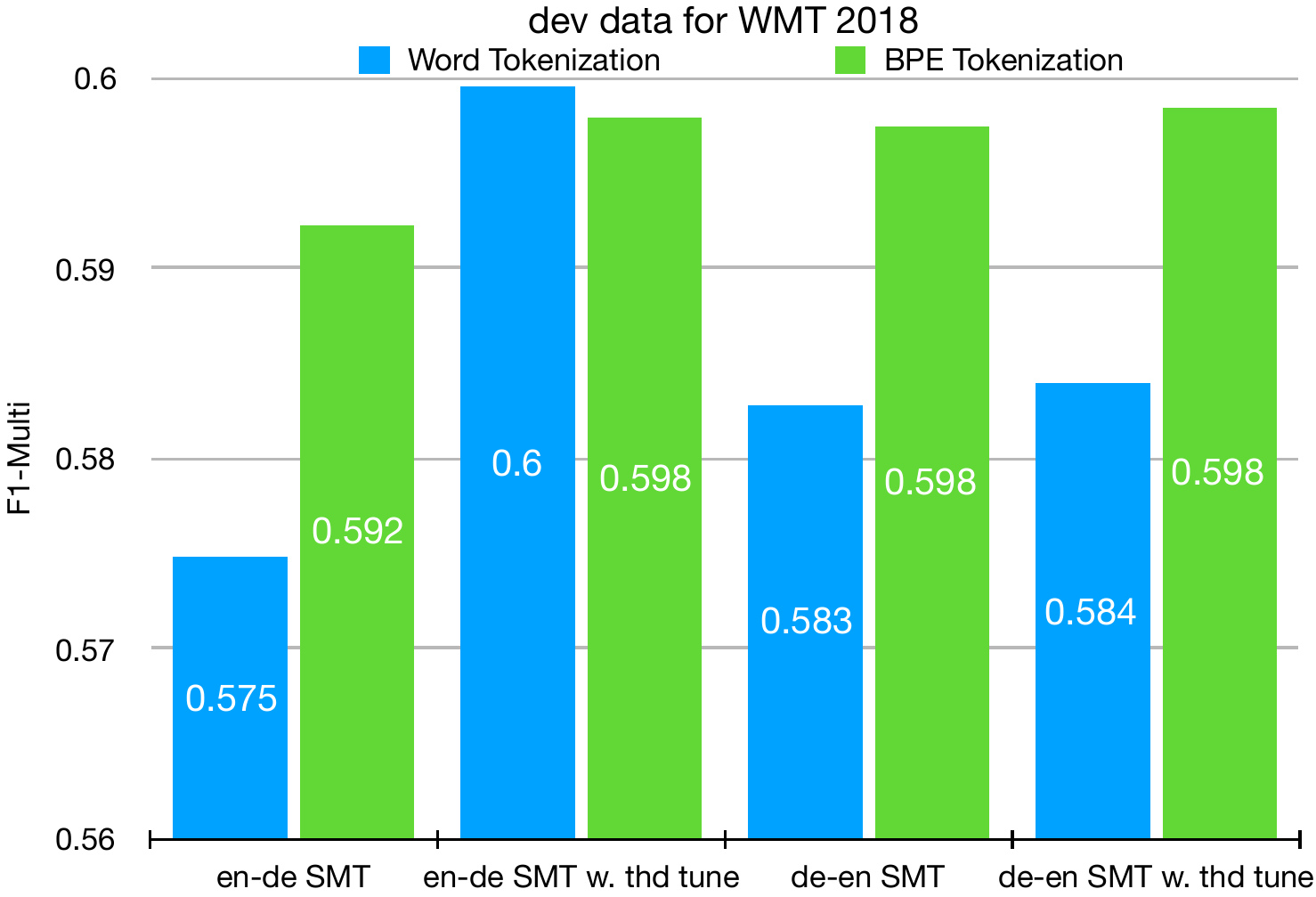}
}
\caption{BPE tokenization results in better results in most experiments.}
\label{fig:BPE}
\end{figure*}

The gap level error prediction is important to machine translation system as well. 
Missing tokens in the machine translation, as indicated by the TER tool, are annotated as follows: after each token in the sentence and at the sentence start, a gap tag is placed. 
Note that the number of gap tags for each translation sentence is $T+1$, including the predictions before the first token and after last one.
Therefore, we can directly build the gap prediction model by modifying (\ref{eq:word_loss}) as,
\begin{equation}\label{eq:gap_loss}
\arg\min \sum_{k=0}^{T} \text{XENT}(g_k, \mathbf{W}[\overrightarrow{\mathbf{h}_k}, \overleftarrow{\mathbf{h}_k},\overrightarrow{\mathbf{h}_{k+1}}, \overleftarrow{\mathbf{h}_{k+1}}])
\end{equation}
where $g_k$ is the gap tag between the $k$th and $k$+1st tokens. 
We can train the neural bilingual expert model for gap prediction to extract more representative features for the downstream task. 
Basically, we have the following factorization model $p(\mathbf{t},\mathbf{t}^g|\mathbf{z})=p(\mathbf{t}|\mathbf{z})p(\mathbf{t}^g|\mathbf{z})$ and $q(\mathbf{z}|\mathbf{s},\mathbf{m})$, where $p(\mathbf{t}|\mathbf{z})$ is identical as previously discussed model, gap token prediction distribution $p(\mathbf{t}^g|\mathbf{z})=\prod_k  p(t_k^g|\overrightarrow{\mathbf{z}_k}, \overleftarrow{\mathbf{z}_k},\overrightarrow{\mathbf{z}_{k+1}}, \overleftarrow{\mathbf{z}_{k+1}})$ and $q$ becomes conditional on $\mathbf{m}$. 
Note that we need to define a ``$<$blank$>$" token for gap prediction, meaning that nothing needs to be inserted.  
Therefore, it also results in a side product -- automatic post-editing. 
If we label the human post-edited translations by the insertion or deletion operations to machine translations (which could be done by using TER tool), we can train the model to predict such operations on the target side, achieving a better APE system eventually. 
We leave this as the future work.

As we discussed in the introduction, most computer assisted translation scenarios use the quality estimation model as the an activator of APE, a guidance to APE corrections, or a selector of final translation output \cite{chatterjee2018combining}. 
Though QE can play the role of a helper function for APE, they are fundamentally considered as two separated tasks. 
In our proposed model, after we pre-trained the neural bilingual model for gap prediction, we can subsequently feed the model derived and mis-matching features to the Bi-LSTM model for gap quality estimation. 
We propose a direction to unify the quality estimation and automatic post-editing. 
First, we demonstrate the performance of our result for gap quality estimation in the left-side of Table~\ref{tab:apegap}.
We also show several examples of APE results by our pre-trained model in the right-side of Table~\ref{tab:apegap}.  

\subsection{Extending to BPE Tokenization} 
In many NMT systems, using BPE or subword units gives an effective way to deal with rare words. 
Especially in German, there are a bunch of compound words, which are simply a combination of two or more words that function as a single unit of meaning, e.g. ``handschuh" means glove in German, which is literally the ``hand shoe".
BPE tokenization gives a good balance between the flexibility of single characters and the efficiency of full words for decoding, and also sidesteps the need for special treatment of unknown words. 

For sentence level HTER prediction, there is no harm or conflict to use BPE, since the regression signals only care about the hidden states of the last time stamps. 
However, for word level labeling, the length of sequential features $L_b$ with BPE tokenization is different from the number of word tokens $L_w$. 
We propose to average the features of all subword units belonging to one single word token, similar to average pooling along the time axis with dynamic sizes. 
To make the computational graph differentiable, the BPE segmentation information needs to be stored into a $L_w \times L_b$ sparse matrix $S$, where $S_{ij}\neq 0$ if $j$-th subword unit belongs to $i$-th word (see Fig~\ref{fig:BPE}(a) for an example). 
The averaged features can be computed by matrix multiplication. 

We compared the performance of the word and BPE tokenization on both sentence and word levels, and results are plotted as histograms in Fig~\ref{fig:BPE}(b,c).
Similar to NMT systems, the finer grained BPE tokenization can improve the QE performance in most tasks. 
In the sentence level, BPE model got a lower Pearson's $r$ for en-de NMT QE task, which is very likely due to the small data size ($<$14000). 
In the word level, if we did not tune the threshold by using the default 0.5, the BPE model can always be better. 
After threshold tuning, the BPE model may have less improvement (we tune the threshold on development data and evaluate on it as well, since we did not have the ground truth of the testing data). 

Actually, the two models can be jointly trained during the stage of quality estimation, no matter the preprocessing is word or BPE tokenization. 
Even for BPE tokenization, we can do back-propagation to update the ``bilingual expert" model when we are training Bi-LSTM, if appropriate column and row paddings are added to the segmentation matrix. 
We will also leave this as another future work. 

\section{Conclusion}

In this paper, we present a novel approach to solve the quality estimation problem for machine translation systems. 
We first introduce the neural ``bilingual expert" model as the prior knowledge model. 
Then, we use a simple Bi-LSTM as the quality estimation model with the extracted model derived and manually designed mis-matching features. 
In the end, we test our algorithm on the public available WMT 17/18 QE competition dataset and yield better performance than other algorithms in most downstream tasks. 

\newpage

\bibliography{aaai}
\bibliographystyle{aaai}

\end{document}